\documentclass{article}
\usepackage{spconf,amsmath,graphicx}
\usepackage{graphicx}
\usepackage{amsmath}
\usepackage{amsthm}
\usepackage{amsfonts}
\usepackage{multirow}
\usepackage{multicol}
\usepackage{array}
\usepackage{booktabs}
\usepackage{color}
\usepackage{tcolorbox}
\usepackage{url}

\newcommand{\figref}[1]{Fig. \ref{#1}}
\newcommand{\tbref}[1]{Table \ref{#1}}
\renewcommand{\eqref}[1]{Eqn. \ref{#1}}

\title{
  Multi-grained Label Refinement Network with Dependency Structures for Joint
  Intent Detection and Slot Filling
}
\name{Baohang Zhou, Ying Zhang\sthanks{Corresponding author.}, Xuhui Sui, Kehui Song, Xiaojie Yuan}
\address{College of Computer Science, TKLNDST, Nankai University, China}
\begin{document}
%\ninept
%
\maketitle
\begin{abstract}
  Slot filling and intent detection are two fundamental tasks in the field of
  natural language understanding.
  Due to the strong correlation between these two tasks, previous studies make
  efforts on modeling them with multi-task learning or designing feature
  interaction modules to improve the performance of each task.
  However, none of the existing approaches consider the relevance between the
  structural information of sentences and the label semantics of two tasks.
  The intent and semantic components of a utterance are dependent on the
  syntactic elements of a sentence.
  In this paper, we investigate a multi-grained label refinement network, which
  utilizes dependency structures and label semantic embeddings.
  Considering to enhance syntactic representations, we introduce the dependency
  structures of sentences into our model by graph attention layer.
  To capture the semantic dependency between the syntactic information and task
  labels, we combine the task specific features with corresponding label
  embeddings by attention mechanism.
  The experimental results demonstrate that our model achieves the competitive
  performance on two public datasets.
\end{abstract}

\section{Introduction}
Slot filling and intent detection are two critical tasks for natural language
understanding (NLU).
The two tasks are defined to identify intents and extract semantic components
from utterances in dialog systems~\cite{DBLP:journals/corr/abs-2101-08091}.
Intent detection is formulated as a sentence-level classification problem, and
slot filling can be regarded as a sequence labeling problem.
Traditional methods tend to solve the two tasks independently.
For intent detection, researchers applied traditional machine learning methods,
such as logistic regression, random forest, and deep belief
networks~\cite{tur2011spoken}.
The sequence-based models, such as long short-term memory
(LSTM)~\cite{DBLP:journals/neco/HochreiterS97} and conditional random fields
(CRF)~\cite{DBLP:conf/icml/LaffertyMP01}, have achieved significant performances
on slot filling.

Considering the relevance between intent detection and slot filling, some works
proposed joint models to tackle the two tasks with feature interaction between
them.
Previous joint learning methods utilized the supervised signal from intent
detection to improve the performance of slot filling by
attention~\cite{DBLP:conf/interspeech/LiuL16} or
gated~\cite{DBLP:conf/naacl/GooGHHCHC18} mechanisms.
Recently, Qin et al.~\cite{DBLP:conf/icassp/QinLCKZ021} proposed a
co-interactive module to model the cross-impact of two tasks and achieved the
state-of-the-art performance.
The joint learning methods always demonstrate their effectiveness over the
independent
models~\cite{DBLP:conf/interspeech/Hakkani-TurTCCG16,DBLP:conf/acl/ZhangLDFY19,DBLP:conf/icassp/HuiWCYWX21}.
However, the existing methods do not take the advantage of the correlation
between the syntactic information and target label semantics.
The syntactic information implies the dependency structures of sentences.
And the objects of prepositional phrases are often slot values to be extracted
in a sentence while the intent of speakers can be reflected by the verbs.
Therefore, it is a meaningful way to exploit the dependency structures of
syntactic for enhancing the sentence representations cooperated with label
semantics.

To address the limitations of existing approaches, we propose a multi-grained
label refinement network with dependency structures for jointly modeling slot
filling and intent detection.
When utilizing the syntactic knowledge, Wang et
al.~\cite{DBLP:conf/aaai/WangWRZC21} proposed an task to predict the dependency
matrix.
Considering the variant importance of syntactic characteristics in dependency
structures, we encode the syntactic information by graph attention
network~\cite{velickovic2018graph}.
Furthermore, we acquire the semantic embeddings of task labels by their
descriptions~\cite{DBLP:journals/taslp/ZhuZMY20}.
Through the attention mechanism~\cite{zhou-etal-2016-attention}, we fuse the
syntactic-enhanced sentence features with prior label semantics of slot filling
and intent detection.
The above operations can bridge the gap between the syntactic information and
label semantics, improving the performance of the two tasks.

We compare our model with several state-of-the-art baselines on two public
datasets: ATIS~\cite{DBLP:conf/naacl/HemphillGD90} and
SNIPS~\cite{DBLP:journals/corr/abs-1805-10190}.
The experimental results demonstrate that our model can achieve significant
improvements on different metrics.
%
% The code is available at: \url{https://github.com/zhoubaohang/SF-ID}.

\section{Model}
\begin{figure*}
    \centering
    \includegraphics[width=0.85\textwidth]{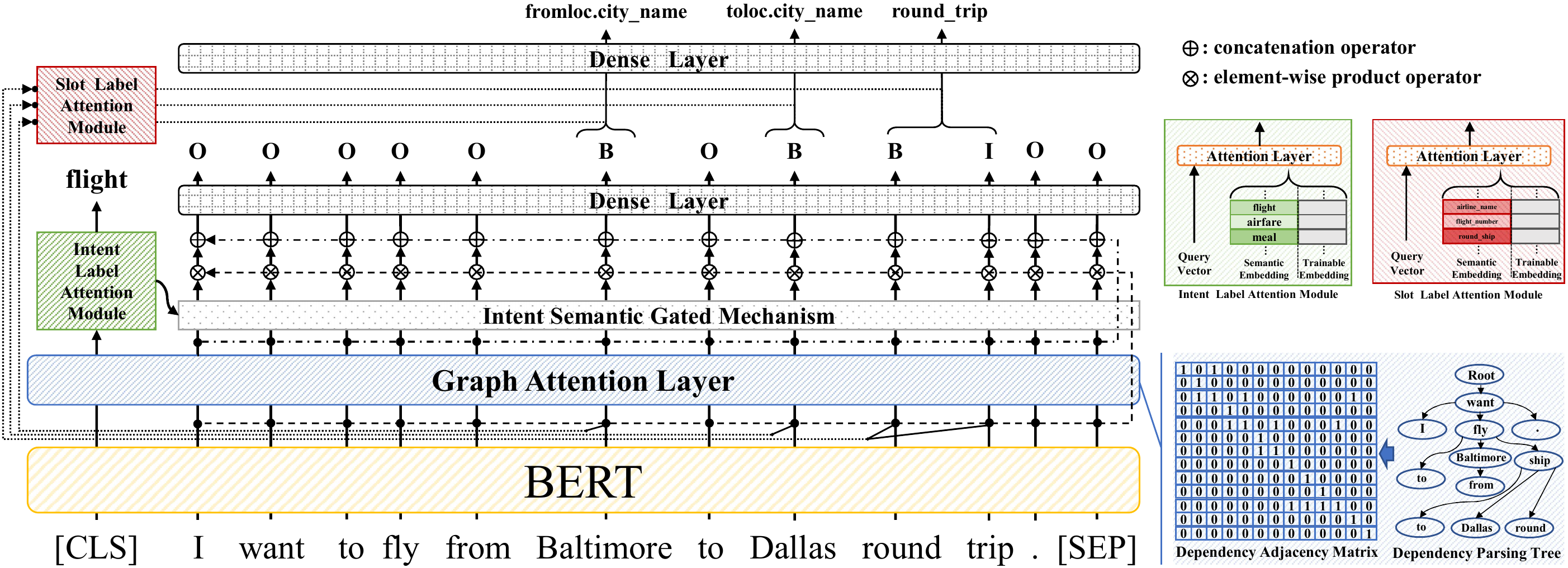}
    \caption{
        The multi-grained label refinement network with dependency structures.
    }
    \label{fig:model}
\end{figure*}

The overall model is shown in \figref{fig:model}.
Before presenting the details of our model, we introduce the notations about the
slot filling and intent detection.
For slot filling, we denote $\{(\textbf{X}_i, \textbf{y}^S_i)\}_{i=1}^{N_s}$ as
a training set with $N_s$ samples, where $\textbf{X}_i$ is the utterance text and
$\textbf{y}_i^S$ is slot filling label.
Given a sentence with $N_w$ words, the utterance text can be formulated as
$\textbf{X} = \{\textbf{x}_i\}_{i=1}^{N_w}$ and the slot filling label is
$\textbf{y}^S = \{y_i\}_{i=1}^{N_w}$.
The location of slot entity is more related to the syntactic of sentence and the
type of slot entity is more related to the semantic of sentence.
Therefore, in our model, we split the slot filling labels $\textbf{y}^S$ into
two parts: \texttt{BIO} labels $\textbf{y}^{ss} = \{y_i^{ss}\}_{i=1}^{N_w}$ and
slot entity types $\{(l_k,r_k,y_k^{st}) | 1 \leq l_k \leq r_k \leq N_w,
    k\in[1,N_e] \}$, where $l_k$ and $r_k$ are the start and end indexes of $k$th
slot entity and $y_k^{st}$ is the type of it in sentence.
For example, the original label sequence ``O B-round\_trip I-round\_trip'' is
reformulated as ``O B I'' and $\{(2,3,\text{``round\_trip''})\}$.
For slot filling, we should predict \texttt{BIO} labels to locate slot entity
first and then acquire the types of predicted slot entities.
Besides, we denote $\{(\textbf{X}_i, y_i^I)\}_{i=1}^{N_s}$ as a intent detection
training set.

\subsection{Pre-trained Language Model}
To map the discrete words of sentences into the dense distributed
representations, we utilize the pre-trained language model
BERT~\cite{DBLP:conf/naacl/DevlinCLT19} as feature extractor.
The BERT architecture can capture the contextual features of sentences
effectively.
Given the input sentence $\textbf{X}$, we should insert special tokens
$[\texttt{CLS}]$ and $[\texttt{SEP}]$ into the start and end of the sentence.
And the feature extraction process can be simplified as
$\text{BERT}(\textbf{X})=\{\textbf{h}_i\}_{i=0}^{N_w+1}$, where $\textbf{h}_0$
is the dense vector of token $[\texttt{CLS}]$, $\{\textbf{h}\}_{i=1}^{N_w}$ is
the language representations of input sentence and $\textbf{h}_i \in
    \mathbb{R}^{d}$.
$d$ is the dimension number of vector extracted from BERT.

\subsection{Dependency Structures Encoder}
The most important contribution of our model is that we exploit the syntactic
structural information to enhance the representations of sentences.
Given the input sentence, we utilize the Stanford CoreNLP toolkit to acquire the
syntactic structures.
According to the direct information of dependency parsing tree, we construct a
adjacency matrix $\textbf{A} \in \mathbb{R}^{(N_w+2) \times (N_w+2)}$ as shown
in \figref{fig:model}.
To capture the graph structural information, we make efforts on the graph neural
networks to encode the syntactic knowledge into our model.
Furthermore, we consider the importance of various syntactic elements and
utilize graph attention network (GAT)~\cite{velickovic2018graph} to enhance the
representations of sentences.
After acquiring the features $\textbf{H} = \{\textbf{h}_i\}_{i=0}^{N_w+1}$, we
feed them into a typical GAT and the syntactic-enhanced feature is $\textbf{G} =
    \{\textbf{g}_i\}_{i=0}^{N_w+1}$.
The calculation process of GAT can be simplified as $\textbf{g}_i =
    \lVert_{k=1}^K \sigma\left(\sum_{j \in n[i]} \alpha_{ij}^k \textbf{W}_G^k
    \textbf{h}_j \right)$, $\alpha_{ij}^k =
    \frac{\exp\left(\text{LeakyRelu}(\textbf{a}_k^T [\textbf{W}_G^k \textbf{h}_i
                \lVert \textbf{W}_G^k \textbf{h}_j] )\right)}{\sum_{c \in n[i]} \exp
        \left(\text{LeakyRelu} (\textbf{a}_k^T [\textbf{W}_G^k \textbf{h}_i \lVert
                \textbf{W}_G^k \textbf{h}_c]) \right)}$, where $\lVert$ is the concatenation
of vectors, $\textbf{W}_G^k$ and $\textbf{a}_k$ are trainable weights of
$k$th multi-head attention, $n[i]$ is the neighbors of input feature
$\textbf{h}_i$ according to the adjacency matrix $\textbf{A}$.

\subsection{Intent Detection Procedure}
The token $[\texttt{CLS}]$ is the aggregation placeholder and its corresponding
feature vector $\textbf{g}_0$ can be utilized to predict the intent type
directly~\cite{DBLP:journals/corr/abs-1902-10909}.
Considering label semantics are helpful to refine the intent
labels~\cite{DBLP:conf/emnlp/CuiZ19}, we construct the intent label embedding
$\textbf{E}^I = \{\textbf{e}_i^I\}_{i=1}^{|I_{label}|}$, where $|I_{label}|$ is
the number of intent label.
The intent label embedding is composed by the description semantic embedding
$\textbf{E}^I_S \in \mathbb{R}^{|I_{label}| \times d}$ and global semantic
embedding $\textbf{W}^I_S \in \mathbb{R}^{|I_{label}| \times d}$, where the
former is fixed and the latter is trainable.
We feed each label description into BERT and concatenate the compressed feature
vectors of label descriptions as $\textbf{E}^I_S$.
Besides, to capture the global semantic of intents, we initialize the trainable
weights as global semantic embedding $\textbf{W}^I_S$.
To focus on the most correlated label semantics, we utilize the attention
mechanism~\cite{zhou-etal-2016-attention} to fuse the syntactic-enhanced
features with intent label embedding.
The attention weighted features can be calculated as: $\textbf{h}^I =
    \text{Attention} (\textbf{g}_0, \textbf{E}^I; \textbf{W}_a, \textbf{b}_a) =
    \sum_{i=1}^{|I_{label}|} \beta_i \textbf{e}_i^I$, where the attention score
is defined as $\beta_i = \frac{\exp\left(\textbf{W}_a [\textbf{e}_i^I \lVert
        \textbf{g}_0] + \textbf{b}_a\right)}{\sum_{j=1}^{|I_{label}|} \exp
    \left(\textbf{W}_a [\textbf{e}_j^I \lVert \textbf{g}_0] +
    \textbf{b}_a\right)}$.
$\textbf{W}_a$ and $\textbf{b}_a$ are trainable weights in the intent label
attention module.
We can calculate the prediction probabilities $\hat{y}^I =
    \text{softmax}\left(\textbf{W}_I [\textbf{h}^I \lVert \textbf{g}_0]
    +\textbf{b}_I \right)$.
The loss function for intent detection is formulated as the cross-entropy:
$\mathcal{L}_{intent} = - \sum_{i=1}^{N_s} y_i^I \log \hat{y}_i^I $.

\subsection{Slot Filling Procedure}
In our model, there are two steps to perform slot filling.
The first step is to predict \texttt{BIO} labels for extracting slot entities.
In the second step, we predict the types of the above extracted slot entities.
The intent label-enhanced feature is focused on the tokens correlated to
sentence intents and helpful to extract slot entities.
Therefore, we propose the intent semantic gated mechanism to guide the model to
locate slot entities.
The formulation of the gated mechanism is $\text{Gate}(\textbf{H}^I, \textbf{G})
    = \sigma \left(\textbf{W}_g [\textbf{H}^I \lVert \textbf{G}] +
    \textbf{b}_g\right)$ where $\sigma$ is the sigmoid function and
$\textbf{H}^I = [\textbf{h}^I;\dots; \textbf{h}^I] \in \mathbb{R}^{d \times
        (N_w + 2)}$.
Considering the semantic difference of the syntactic-enhanced features and raw
language ones, we combine the two parts and the fusion sentence feature is
$\textbf{H}^S = [\text{Gate}(\textbf{H}^I, \textbf{G}) \odot \textbf{G} \lVert
    \textbf{H}] $ where $\odot$ is the element-wise production.
We utilize the fusion feature to predict the \texttt{BIO} label probabilities
$\hat{y}_i^{ss} = \text{softmax}\left(\textbf{W}_{o1} \textbf{h}_i^S + \textbf{b}_{o1}
    \right)$.
The loss function of this step is defined as $\mathcal{L}_{slot1} = -
    \sum_{i=1}^{N_s} \sum_{j=1}^{N_w} y_{ij}^{ss} \log \hat{y}_{ij}^{ss}$ in
which $\hat{y}_{ij}^{ss}$ is the \texttt{BIO} label probabilities of the
$j$th token in the $i$th sample.
To predict the type of each slot entity, we utilize the raw language features of
spans as representations: $\textbf{r}_k = \sum_{i=l_k}^{r_k} \textbf{h}_i$.
To utilize the label semantics for refining the slot labels, we also construct
the slot label embedding $\textbf{E}^S = \{\textbf{e}_i^S\}_{i=1}^{|S_{label}|}$
as the way for intent label embedding.
The label semantic features are fused with language features as $\textbf{h}^S
    = \text{Attention}(\textbf{r}, \textbf{E}^S; \textbf{W}_v, \textbf{b}_v)$.
We predict the type of slot entity as $\hat{y}^{st} = \text{softmax}
    \left(\textbf{W}_{o2} \textbf{h}^S + \textbf{b}_{o2}\right)$.
The loss function of the second step is $\mathcal{L}_{slot2} = -
    \sum_{i=1}^{N_s} \sum_{k=1}^{N_e} y_{ik}^{st} \log \hat{y}_{ik}^{st}$ where
$\hat{y}_{ik}^{st}$ is the type probabilities of the $k$th slot entity in
the $i$th sample.

\subsection{Training Procedure}
To tackle the slot filling and intent detection tasks at once, we introduce the
hyper-parameter to sum the loss functions $\mathcal{L}_{intent}$,
$\mathcal{L}_{slot1}$ and $\mathcal{L}_{slot2}$.
The overall loss function for the joint learning model is defined as:
$\mathcal{L} = \left(1 - \gamma\right) \cdot \left(\mathcal{L}_{slot1} +
\mathcal{L}_{slot2}\right) + \gamma \cdot \mathcal{L}_{intent}$ where $\gamma$
is the hyper-parameter for balancing different task losses.
For training the model, we feed the training samples into it and calculate the
overall loss by the above equation.
And then we utilize the stochastic gradient descent method to update the
parameters of the model.

\section{Experiments}
We compare our model with baseline methods on two public datasets.
\begin{table}[ht]
    \center
    \begin{tabular}{lcc}
        \toprule
        Hyper-parameter                    & ATIS & SNIPS \\
        \midrule
        $\gamma$                           & 0.6  & 0.5   \\
        batch size                         & 16   & 14    \\
        learning rate                      & 1e-5 & 1e-5  \\
        \# graph attention head            & 4    & 2     \\
        graph attention dropout rate       & 0.4  & 0.5   \\
        \# graph attention output features & 256  & 512   \\
        \bottomrule
    \end{tabular}
    \caption{
        The hyper-parameter settings for our model on the ATIS and SNIPS
        datasets.
    }
    \label{tb:settings}
\end{table}
The airline travel information systems (ATIS) dataset contains recordings of
people having flight services~\cite{DBLP:conf/naacl/HemphillGD90}.
The SNIPS dataset covers utterances of different domains, such as: weather,
restaurants and entertainment~\cite{DBLP:journals/corr/abs-1805-10190}.
Both two original datasets are not split into training, validation and test
sets.
We follow the same format and partition of datasets as in Qin et
al.~\cite{DBLP:conf/emnlp/QinCLWL19}.

The hyper-parameter settings of our model on two dataset are shown in
\tbref{tb:settings}.
And we use Adam~\cite{DBLP:journals/corr/KingmaB14} algorithm to optimize the
trainable parameters in our model.
To compare with different models, we evaluate performance of slot filling using
entity-level F1 score, intent detection using accuracy and sentence-level
semantic parsing using overall accuracy.
We save the best model which achieves the highest score on the validation set
and report the results of it on the test set.

In this paper, we utilize the \texttt{BERT-large} version of pre-trained model
BERT as language model.
To demonstrate the effectiveness of the proposed method fairly, we compare our
model with BERT-based models, such as
Stack-Propagation~\cite{DBLP:conf/emnlp/QinCLWL19},
BERT-joint~\cite{DBLP:journals/corr/abs-1902-10909},
SlotRefine~\cite{DBLP:conf/emnlp/WuDLX20} and
SyntacticTF~\cite{DBLP:conf/aaai/WangWRZC21}.
Besides, we also select LSTM-based models as baselines to show the superiority
of pre-trained language model and the effectiveness of our model.

\subsection{Experimental Results}
The detailed experiment results on ATIS and SNIPS are shown in
\tbref{tb:main-results}.
Conventional joint learning models, such as
Slot-Gated~\cite{DBLP:conf/naacl/GooGHHCHC18} and
CapsuleNLU~\cite{DBLP:conf/acl/ZhangLDFY19}, utilized the intent information to
improve the performance of slot filling.
Wu et al. proposed SlotRefine with two-stage training process for refining
intent and slot labels~\cite{DBLP:conf/emnlp/WuDLX20}.
SyntacticTF~\cite{DBLP:conf/aaai/WangWRZC21} introduced dependency parsing
prediction task into slot filling and intent detection joint learning model, and
achieved the state-of-the-art results.
Compared with existing methods, we bridge the gap between the syntactic
information and task label semantics.
While utilizing syntactic knowledge to enhance the representations of sentences,
we also exploit the slot-level and intent-level label semantic features with
attention mechanism to improve the performance of two tasks.
\begin{table*}[ht]
    \center
    \resizebox{0.85\textwidth}{!}{
        \begin{tabular}{l|ccc|ccc}
            \hline
            \multirow{2}{*}{Model}                                   & \multicolumn{3}{c|}{ATIS} & \multicolumn{3}{c}{SNIPS}                                                                     \\ \cline{2-7}
                                                                     & Slot~(F1)                 & Intent~(Acc)              & Semantic~(Acc) & Slot~(F1)      & Intent~(Acc)   & Semantic~(Acc) \\ \hline
            Attention-based RNN~\cite{DBLP:conf/interspeech/LiuL16}  & 94.20                     & 91.10                     & 78.90          & 87.80          & 96.70          & 74.10          \\
            Joint Seq~\cite{DBLP:conf/interspeech/Hakkani-TurTCCG16} & 94.30                     & 92.60                     & 80.70          & 87.30          & 96.90          & 73.20          \\
            Slot-Gated~\cite{DBLP:conf/naacl/GooGHHCHC18}            & 95.20                     & 94.10                     & 82.60          & 88.30          & 96.80          & 74.60          \\
            CapsuleNLU~\cite{DBLP:conf/acl/ZhangLDFY19}              & 95.20                     & 95.00                     & 83.40          & 91.80          & 97.30          & 80.90          \\
            BiLSTM-CRF~\cite{DBLP:conf/semco/DahaH19}                & 95.60                     & 96.60                     & 86.20          & 94.60          & 97.40          & 87.20          \\
            ELMo~\cite{DBLP:conf/aaai/SiddhantGM19}                  & 95.62                     & 97.42                     & 87.35          & 93.90          & 99.29          & 85.43          \\
            BERT-Joint~\cite{DBLP:journals/corr/abs-1902-10909}      & 96.10                     & 97.50                     & 88.20          & 97.00          & 98.60          & 92.80          \\
            Stack-Propagation~\cite{DBLP:conf/emnlp/QinCLWL19}       & 96.10                     & 97.50                     & 88.60          & 97.00          & 99.00          & 92.90          \\
            GraphLSTM~\cite{DBLP:conf/aaai/ZhangMZYW20}              & 95.91                     & 97.20                     & -              & 95.30          & 98.29          & -              \\
            SlotRefine~\cite{DBLP:conf/emnlp/WuDLX20}                & 96.16                     & 97.74                     & 88.64          & 97.05          & 99.04          & 92.96          \\
            SyntacticTF~\cite{DBLP:conf/aaai/WangWRZC21}             & 96.01                     & 97.31                     & -              & 96.89          & 99.14          & -              \\ \hline
            Ours                                                     & \textbf{96.28}            & \textbf{98.78}            & \textbf{89.79} & \textbf{97.17} & \textbf{98.51} & \textbf{93.26} \\
            w/o slot label attention module                          & 95.89                     & 98.50                     & 88.54          & 96.50          & 98.32          & 92.10          \\
            w/o intent label attention module                        & 96.10                     & 98.10                     & 88.42          & 96.78          & 98.10          & 92.25          \\
            w/o dependency structures encoder                        & 96.03                     & 98.17                     & 88.40          & 96.36          & 98.14          & 91.24          \\\hline
        \end{tabular}
    }
    \caption{
        Performances of different models on two datasets.
        The results of ablation study for our model are also presented.
    }
    \label{tb:main-results}
\end{table*}

Our model does not only achieve competitive scores on unilateral evaluation
metrics but also gain significant improvements on the integrated metric over
baseline methods.
Especially on semantic accuracy, our model exceeds the best baselines by 1.3\%
and 0.3\% on ATIS and SNIPS respectively.
The improvement verifies that the proposed model further establishes the
semantic association between slot filling and intent detection.
Compared with traditional models which do not utilize label semantics, our model
achieves higher scores on two tasks.
Although BERT-joint~\cite{DBLP:journals/corr/abs-1902-10909} utilized BERT to
jointly model slot filling and intent detection, our model takes advantage of
pre-trained language and rich prior knowledge including: label semantics and
syntactic information.
Therefore, the proposed model outperforms the BERT-based models on almost
metrics.
Our model does not achieve the best result of intent accuracy on SNIPS dataset.
There might be complicated domains included in the dataset that our model cannot
handle.

\subsection{Ablation Study}
\noindent \textbf{Impact of Slot Label Attention Module} We remove the slot
label attention module and directly use the span features $\textbf{r}$ to
predict the slot entity types.
The results of experiment ``w/o slot label attention module'' are presented in
\tbref{tb:main-results}.
We can observe that the slot filling performance decreased the most, which
demonstrates the slot label semantics are critical to the slot filling
procedure.
Besides, the intent detection performance also drops a little, and the implicit
correlation between intent detection and slot filling influences the
performances of each other.

\noindent \textbf{Impact of Intent Label Attention Module} After removing the
intent label attention module, we feed the language representation
$\textbf{h}_0$ of token \texttt{[CLS]} into the intent semantic gated mechanism
and use it to predict the intent labels.
The results of experiment ``w/o intent label attention module'' are shown in
\tbref{tb:main-results}.
We can observe that the performance of intent detection decreased the most while
the slot filling performance also declined to a certain extent.
This phenomenon verifies that intent label semantics play an important role in
the intent detection and are beneficial to improving the slot filling
performance.

\noindent \textbf{Impact of Dependency Structures Encoder}
To prove the effectiveness of syntactic knowledge, we conduct the experiment
``w/o dependency structures encoder'' as shown in \tbref{tb:main-results} and
utilize language representations $\textbf{H}$ to perform slot filling and intent
detection.
We can observe that the slot filling and intent detection performances both
declined a lot.
This proves that the prior syntactic knowledge is critical to the two tasks.
The dependency structural information is encoded by GAT and enhances the
representations of sentences to improve the performance of slot filling and
intent detection.

\section{Conclusion}
In this paper, we propose a multi-grained label refinement network with
dependency structures.
Considering the implicit correlation between syntactic knowledge and task label
semantics, we encode the dependency structural information by graph attention
network, and utilize slot and intent label attention modules to fuse the
syntactic-enhanced features with label semantic ones.
Experimental results on two benchmarks demonstrate the superiority of our model.
In the future, we will tackle cross-domain slot filling and intent detection
with prior knowledge driven models.

\section{Acknowledgements}
This research is supported by the Chinese Scientific and Technical Innovation
Project 2030 (2018AAA0102100), NSFC-General Technology Joint Fund for Basic
Research (No. U1936206), NSFC-Xinjiang Joint Fund (No. U1903128), National
Natural Science Foundation of China (No. 62002178, No. 62077031), and Natural
Science Foundation of Tianjin, China (No. 20JCQNJC01730).

\vfill\pagebreak
\bibliographystyle{IEEEbib}
\bibliography{refs}

\end{document}